\documentclass[sigconf]{acmart}

\usepackage{booktabs}
\usepackage{multirow}
\usepackage{graphicx}
\usepackage{subcaption}

\newcommand{\floatnote}[1]{\par\vspace{2pt}\footnotesize\emph{Note.} #1}

\renewcommand\footnotetextcopyrightpermission[1]{}
\acmConference[AIDataSci '26]{The 1st International Workshop on AI Data Scientist}{2026}{Jeju, Korea}

\begin{document}

\title{Do LLM-Generated Skills Make Better AI Data Scientists? A Component Ablation Across Data-Science Workflows}

\author{Wei-Jung Huang}
\affiliation{%
  \institution{Independent Researcher}
  \country{United States}
}

\begin{abstract}
Product data scientists often ask LLM-based agents to help with recurring execution tasks such as cleaning data, writing SQL, choosing statistical tests, and formatting results. Reusable skill files are meant to avoid prompting from scratch by packaging guidance for a task family. Expert-written skills can encode high-quality guidance, but writing and maintaining them across many data-science task families creates a manual bottleneck. We ask whether LLM-generated skills offer a useful low-curation alternative: do they improve performance over the task prompt alone?

We test this question across four lifecycle stages: data preparation, data extraction, statistical analysis, and reporting, using one generated skill per stage. We find no reliable improvement from full generated skills over No-Skill prompting. We then ask whether any part of the skill is useful by ablating different skill components. The main ablation covers 56 tasks, nine model configurations, and three providers, yielding 7{,}560 runs.

Compared with prompting using the task alone, neither the full generated skill nor any ablated skill variant significantly improves performance; all $p$-values are at least 0.396, and the total spread across variants is only 1.2~pp. A supplemental token-matched control adds 1{,}512 runs and finds that Full skills perform similarly to task-irrelevant skill-formatted content. The results caution against using one LLM-generated skill per data-science workflow as a default single-shot prompting strategy.
\end{abstract}



\maketitle

\section{Introduction}

Product data scientists often ask LLM-based agents to help with recurring execution tasks: cleaning data, writing SQL, choosing statistical tests, computing effect sizes, and formatting reports. We refer to these systems as data-science agents, following recent benchmarks for data-science automation~\cite{datascibench2025, ds1000}. For these systems, model choice is only part of the design problem. The other part is how to supply domain knowledge.

Recent agent platforms and benchmarks use reusable skill files, such as \texttt{SKILL.md}, to package task instructions, examples, and reference notes~\cite{anthropicskills2025, skillsbench2026}. For data-science agents, this makes a tempting workflow: write one skill for recurring task families such as data preparation, SQL generation, statistical analysis, and reporting, then prepend it to future tasks.

Expert-written skills can encode high-quality guidance, but they require practitioners to decide what to include, write examples and reference notes, and keep the content current as tools and task conventions change. This manual process scales poorly when teams need guidance for many data-science task families. LLM-generated skills offer a lower-curation alternative: generate task-family guidance once and reuse it across related tasks. The question is whether this low-curation version works in practice.

Existing evidence suggests that this is not guaranteed. SkillsBench~\cite{skillsbench2026} (86 tasks, 7{,}308 trajectories) found that human-curated skills improve performance substantially ($+$16.2~pp), while LLM-generated skills provide no aggregate benefit. However, SkillsBench does not specifically evaluate reusable skills for data-science workflows.

That null result also raises a component-level question: when an LLM-generated skill fails to improve performance, is every component unhelpful, or are useful sections canceled out by harmful ones? For builders of data-science agents, it is useful to ask both whether LLM-generated skills help on data-science workflows and which failure patterns appear when they do not.

We study these questions across four execution-facing data-science lifecycle stages. Each stage is represented by a task family covering a distinct workflow: \textbf{data preparation} (null handling, deduplication, type coercion), \textbf{data extraction} (SQL query formulation), \textbf{statistical analysis} (hypothesis testing, effect sizes), and \textbf{reporting} (structured JSON reporting). These workflows represent common execution tasks for data-science agents, and their outputs can be checked by deterministic verifiers.

Our evaluation has two parts. First, we compare task-only prompting with one full LLM-generated skill per stage. Second, we ablate the same skills to test whether any section helps on its own. We use a simplified single-shot setup so that the comparison focuses on the skill content itself. Section~\ref{sec:methods} describes this design choice.

We make three contributions:
\begin{itemize}
    \item We evaluate low-curation LLM-generated skills for data-science agents across four lifecycle stages, 56 tasks, nine model configurations, and 7{,}560 main-ablation runs.
    \item We ablate procedures, examples, and reference notes, then add a 1{,}512-run token-matched control to separate data-science content from prompt-length overhead.
    \item We observe no reliable improvement from either full skills or ablated variants, and analyze failure patterns that differ across task regimes.
\end{itemize}

\section{Related Work}

Our study sits at the intersection of prompt-level knowledge injection, reusable skills, and LLM benchmarks for data-science tasks.

The \texttt{SKILL.md} specification is one recent form of reusable domain instruction for LLM prompts~\cite{anthropicskills2025}. Related ways of supplying task context include retrieval-augmented generation~\cite{lewis2020rag} and few-shot prompting~\cite{brown2020fewshot}. Prior work also shows that prompt components can affect model behavior in non-obvious ways: \citet{min2022rethinking} found that label correctness in demonstrations matters less than format and input distribution, and \citet{lu2022fantastically} showed that example ordering can significantly affect performance. We move from individual prompt elements to reusable skill files and ask whether their content helps when injected into data-science tasks.

Reusable skills have also been evaluated directly. SkillsBench~\cite{skillsbench2026} found that focused human-curated skills improve performance, while LLM-generated skills provide no aggregate benefit. SkillLearnBench~\cite{skilllearnbench2026} studies continual skill-learning methods and finds gains over no-skill baselines, but no method dominates across tasks and models. SkVM~\cite{skvm2026} treats skills as compiled capabilities across heterogeneous LLM backends. These studies motivate testing reusable skills, but they do not ask which parts of generated skills help or hurt in data-science workflows.

On the benchmark side, DS-1000~\cite{ds1000}, DataSciBench~\cite{datascibench2025}, and LLM4DS~\cite{llm4ds2024} evaluate LLM code generation for data-analysis tasks and show that strong models can solve many data-science tasks from the task prompt alone. That creates a hard test for reusable skills: if task-only prompting already covers many common data-science procedures, skills have less room to help and may still add cost or conflict with task-specific instructions. These benchmarks evaluate data-science capability, but they do not test whether reusable skills improve performance in this domain.

Recent data-preparation and table-reasoning systems also point to a different form of support than static front-loaded instructions. PrepBench evaluates natural-language-driven data preparation and emphasizes ambiguous intents, imperfect real-world data, interactive disambiguation, code generation, and workflow translation~\cite{prepbench2026}. AutoDCWorkflow generates data-cleaning workflows from a raw table and analysis purpose, then evaluates answer, data, and workflow quality~\cite{autodcworkflow2025}. Chain-of-Query uses multi-agent collaboration and clause-by-clause SQL generation for table understanding~\cite{chainofquery2025}. These systems do not evaluate reusable skill files directly, but they suggest one limitation of flat prompt injection: data-science agents may need task-specific orchestration, validation, and feedback.

\section{Experimental Design}
\label{sec:methods}

\subsection{Skill Construction}

We generated one skill per lifecycle stage (four skills total) using a Gemini 2.5 Pro coding agent with extended thinking enabled. The agent received the target workflow, required section structure, and \texttt{SKILL.md} schema, and it could research the target domains and related benchmarks before writing each Full skill. Each skill was produced in one autonomous session with no human editing, candidate selection, or iterative refinement. This is the low-curation workflow we study: generate one skill for a task family, then reuse it as written.

Each skill contains four sections:

\begin{itemize}
    \item \emph{Routing} ($\sim$50 tokens): activation triggers specifying when to apply the skill
    \item \emph{Core Procedure} ($\sim$190 tokens): step-by-step workflow instructions
    \item \emph{Worked Examples} ($\sim$225 tokens): concrete input$\to$output demonstrations
    \item \emph{Reference Notes} ($\sim$225 tokens): supplementary heuristics and conventions
\end{itemize}

These sections correspond to the content types in Anthropic's Agent Skills documentation~\cite{anthropicskills2025}. Unlike Anthropic's recommended architecture, which stores reference materials in separate files loaded on demand, we inject all sections as a single flat file. We make this choice to isolate content: the experiment tests whether generated skill content helps when the model sees it, not whether a separate loader can retrieve the right file at the right time. For this reason, the study should not be read as an evaluation of selective loading or progressive disclosure. The format also matches a common low-curation implementation: a single markdown document produced without explicit tooling for multi-file packages.

The ablation variants are produced by mechanically deleting sections from the Full skill, not by separate generation runs. This matters for causal interpretation: the skill variants differ only in which sections are present. Sections are written to be self-contained, so deletion does not leave dangling cross-references. This design lets us compare the full skill against variants that remove supporting content, while keeping the basic task procedure and generation source fixed.

\subsection{Ablation Conditions}

Our ablation treats Worked Examples and Reference Notes as two independent binary factors, with Routing and Core Procedure held constant. We keep Routing and Core Procedure together because, without activation triggers and procedural steps, the remaining sections lack the context needed to function as a coherent skill. Routing is also redundant in our setup because each skill is only paired with tasks from its matching lifecycle stage. This yields four skill conditions plus a No-Skill baseline:

\begin{enumerate}
    \item \textbf{No-Skill}: task prompt only
    \item \textbf{Core-Only}: Routing + Core Procedure
    \item \textbf{Core+Examples}: Routing + Core Procedure + Worked Examples
    \item \textbf{Core+Refs}: Routing + Core Procedure + Reference Notes
    \item \textbf{Full}: all four sections
\end{enumerate}

Skill content is injected unconditionally as a user-message prefix using a fixed template (\texttt{[SKILL START] ... [SKILL END]}, followed by the task prompt) across all conditions and providers.

The No-Skill baseline contains only the task prompt, with no injected skill wrapper or control text. Thus, the comparison is between the same task prompt alone and the same task prompt with generated skill content prepended.

Length-Control tests a simpler alternative explanation: the model may react to a long, skill-formatted prefix rather than to useful data-science guidance. To test this, we add a supplemental \textbf{Length-Control} condition. For each lifecycle stage, we create a task-irrelevant office-supply setup skill with the same injection wrapper, markdown-style organization, and tokenizer length within 1\% of the corresponding Full skill.

We use coherent office-supply guidance rather than random strings, which could create an unrealistic distraction. The control gives the model plausible procedural instructions, examples, and notes, but none are intended to help with data cleaning, SQL, statistical analysis, or reporting. We treat it as a control, not a harmless placebo, since unrelated instructions can still change model behavior. Comparing Full against Length-Control estimates whether the generated data-science content helps beyond adding similarly long, skill-formatted context.

Generic skill guidance can conflict with the task prompt. If the main problem is priority ambiguity, a short rule telling the model to follow the task over the skill should recover some failures. To test this, we define a supplemental \textbf{Full+Priority} condition: it uses the same \texttt{full.md} skill files as Full, but inserts a user-message directive after \texttt{[SKILL END]} and before the task prompt stating that task instructions override conflicting skill guidance. This is a prompt-level task-over-skill directive, not a system-level instruction hierarchy or retrieval-gated priority mechanism.

These supplemental conditions are narrow checks around the main flat-injection ablation. They do not test selective loading, system-level instruction priority, or schema/tool constraints.

The design also omits an expert-written or task-specific positive-control skill. This keeps the focus on low-curation LLM-generated skills, but it means the study does not test whether this benchmark would detect gains from carefully authored skill content.

\subsection{Tasks and Lifecycle Mapping}

We evaluate 56 tasks (14 per lifecycle stage), mapped to four data-science workflows:

\begin{itemize}
    \item \textbf{Data Preparation} (14 tasks): CSV cleaning, including null imputation, deduplication, type coercion, and date standardization
    \item \textbf{Data Extraction} (14 tasks): SQL query formulation against relational databases, using schemas from Spider~\cite{yu2018spider} and standard benchmark databases
    \item \textbf{Statistical Analysis} (14 tasks): hypothesis testing, effect-size computation, and statistical inference using standard datasets (Fisher Iris~\cite{fisher1936iris}, mtcars, PlantGrowth, ToothGrowth)
    \item \textbf{Reporting} (14 tasks): generating structured JSON reports from API responses (GitHub Events, OpenWeatherMap, REST Countries)
\end{itemize}

These tasks cover the verifiable execution layer of product data-science work rather than the full product decision-making lifecycle. We intentionally exclude upstream product framing and experiment design because those tasks often depend on business context, stakeholder constraints, engineering feasibility, and judgment calls that are not well captured by the deterministic verifiers used here.

Tasks are split evenly between self-authored and externally grounded examples to reduce bias from task authorship. The \textbf{self-authored} subset contains 28 tasks (7 per stage). A Gemini 2.5 Pro coding agent with extended thinking enabled drafted the prompts and initial gold outputs for these tasks.

The \textbf{externally grounded} subset also contains 28 tasks (7 per stage) and uses public datasets, benchmark schemas, or public APIs rather than purely synthetic inputs.\footnote{SQL tasks use schemas from Spider~\cite{yu2018spider} and Chinook/Northwind-style databases; Data Preparation tasks use the Melbourne Housing, Titanic, UCI Wine Quality, Auto MPG, and student performance datasets; Statistical Analysis tasks use Fisher Iris~\cite{fisher1936iris}, mtcars, PlantGrowth, ToothGrowth, and the sleep dataset; Reporting tasks use the GitHub Events, OpenWeatherMap, and REST Countries APIs.}

Gold outputs were then validated by scripts that recompute expected results from source data where applicable. Scoring uses fixed gold files and deterministic verifiers, so Gemini does not act as the evaluator at test time. All task prompts and gold outputs were finalized before skill generation, preventing post-hoc alignment between task requirements and skill content.

\subsection{Verification}

All outcomes are binary pass/fail, evaluated by fully automated, deterministic verifiers with no human judgment involved. Each lifecycle stage uses a verifier matched to its output type:

\begin{itemize}
    \item \textbf{Data Preparation}: the model's output CSV is parsed into a DataFrame and compared against the gold CSV. Exact-mode tasks check row-level values, ignoring column order but preserving case sensitivity. Structural-mode tasks check expected columns, row counts, absence of nulls, and date-format compliance.
    \item \textbf{Data Extraction}: the model's SQL is executed against an in-memory SQLite database seeded with task-specific test data. The result set is compared to the gold query output with order-insensitive row matching.
    \item \textbf{Statistical Analysis}: numeric outputs (test statistics, p-values, effect sizes) are extracted and compared against gold values with tolerance $\epsilon=0.01$. Both absolute and relative tolerance are applied to accommodate rounding differences.
    \item \textbf{Reporting}: JSON output is validated against a task-specific schema, including required keys, value types, nested structure, and selected field values where specified.
\end{itemize}

Exact-match verifiers can conflate reasoning failures with format errors. To check this risk, we programmatically identified all condition-flip cases, where a task passes under one condition but fails under another, and reviewed the associated verifier messages. In these cases, failures reflected substantive errors, such as wrong imputation method or wrong case convention due to competing instructions, rather than trivial formatting mismatches.

\subsection{Models}

We select nine model configurations spanning three providers (Table~\ref{tab:models}), varying along three dimensions. First, \emph{model capability}: we include compact models (GPT-4o-mini, Gemini Flash, Claude Haiku) and frontier default models (GPT-4o, Gemini Pro, Claude Sonnet) to test whether stronger models use skills differently. Second, \emph{provider diversity}: sampling from OpenAI, Google, and Anthropic guards against provider-specific artifacts.

Third, \emph{reasoning mode}: Gemini 2.5 Flash and Claude Sonnet 4 are each tested with extended thinking disabled and enabled, and o3-mini provides an always-on reasoning baseline. For stratified analysis, we use three pre-specified groups: compact default models, frontier default models, and explicit-reasoning configurations. Gemini 2.5 Pro remains in the frontier default group because it was run only in its default configuration, with no paired disabled-thinking condition. Moving it into the explicit-reasoning group would mix model scale with reasoning-mode treatment and break the balanced design.

\begin{table}[ht]
\centering
\small
\begin{tabular}{@{}llll@{}}
\toprule
\textbf{Model} & \textbf{Provider} & \textbf{Reasoning Mode} & \textbf{Group} \\
\midrule
GPT-4o-mini           & OpenAI    & No       & Compact \\
GPT-4o                & OpenAI    & No       & Frontier \\
o3-mini               & OpenAI    & Native   & Explicit-reasoning \\
Gemini 2.5 Flash      & Google    & Disabled & Compact \\
Gemini 2.5 Flash$^+$  & Google    & Enabled  & Explicit-reasoning \\
Gemini 2.5 Pro        & Google    & Default  & Frontier \\
Claude Haiku 4.5      & Anthropic & No       & Compact \\
Claude Sonnet 4       & Anthropic & Disabled & Frontier \\
Claude Sonnet 4$^+$   & Anthropic & Enabled  & Explicit-reasoning \\
\bottomrule
\end{tabular}
\caption{Model configurations.}
\label{tab:models}
\floatnote{Groups are balanced by design. Explicit-reasoning denotes configurations run through explicit reasoning-mode settings or a reasoning-only API; it does not imply that other models lack internal reasoning. Gemini 2.5 Pro is treated as frontier default because there is no paired disabled-thinking Gemini Pro condition. Temperature is set to 0 where supported; o3-mini and Claude Sonnet 4$^+$ use provider-required reasoning settings. Temperature 0 reduces sampling variation but does not make calls deterministic.}
\end{table}

\noindent Exact API model identifiers are recorded in the evaluation harness. The $^+$ variants use the same base identifiers as their non-$^+$ counterparts, with extended thinking enabled.

Each of the 56 $\times$ 5 $\times$ 9 = 2{,}520 main-ablation cells is run three times, yielding 7{,}560 runs. The supplemental Length-Control and Full+Priority conditions each add 56 $\times$ 1 $\times$ 9 $\times$ 3 = 1{,}512 runs, for 10{,}584 total runs. We use three repetitions because API providers can exhibit minor non-determinism even under low-variation decoding settings, due to factors such as batching and quantization. Of the 2{,}520 main-ablation cells, 93.4\% are unanimous (all three repetitions produce the same pass/fail outcome) and only 6.6\% show a split verdict. We use majority vote (2-of-3 pass) as the cell-level outcome to avoid treating repetitions as independent observations.

\subsection{Analysis}

We analyze results with four complementary methods:
\begin{itemize}
    \item \textbf{Mixed-effects model.} We fit a linear mixed-effects model with condition as a fixed effect, task as a random-intercept grouping factor, and model as a random-effect variance component. This estimates condition effects after accounting for task difficulty and model strength. We report the linear model because its coefficients directly express percentage-point differences; a logistic GEE and a non-parametric permutation test give substantively identical conclusions.
    \item \textbf{Group-stratified bootstrap analysis.} We estimate skill effects separately for compact, frontier, and explicit-reasoning groups to test whether effects differ across model configurations.
    \item \textbf{Bootstrap confidence intervals.} We compute percentile bootstrap intervals with 10{,}000 iterations over the 56 task-level paired differences.
    \item \textbf{McNemar tests.} For each skill condition, we count task-model pairs where skill content flips the outcome from fail to pass versus pass to fail, then test whether these counts differ.
\end{itemize}
We analyze Length-Control and Full+Priority as supplemental paired controls against No-Skill and Full, rather than as skill components in the main ablation.

\section{Results}

\subsection{LLM-Generated Skills Do Not Improve Performance}

\begin{table*}[ht]
\centering
\small
\begin{tabular}{@{}l ccccc@{}}
\toprule
& \textbf{No-Skill} & \textbf{Full} & \textbf{Core-Only} & \textbf{Core+Ex} & \textbf{Core+Refs} \\
\midrule
GPT-4o-mini           & 57.1\% & 57.1\% & 57.1\% & \underline{51.8\%} & \textbf{58.9\%} \\
GPT-4o                & \textbf{64.3\%} & \textbf{64.3\%} & \underline{58.9\%} & 60.7\% & 60.7\% \\
o3-mini               & 80.4\% & 80.4\% & \underline{78.6\%} & \textbf{82.1\%} & \textbf{82.1\%} \\
Gemini 2.5 Flash      & \underline{55.4\%} & 58.9\% & 58.9\% & \textbf{62.5\%} & 57.1\% \\
Gemini 2.5 Flash$^+$  & 58.9\% & \underline{57.1\%} & \textbf{60.7\%} & 58.9\% & \underline{57.1\%} \\
Gemini 2.5 Pro        & 69.6\% & \underline{67.9\%} & 71.4\% & \textbf{76.8\%} & 69.6\% \\
Claude Haiku 4.5      & \textbf{67.9\%} & \underline{66.1\%} & \underline{66.1\%} & \underline{66.1\%} & \underline{66.1\%} \\
Claude Sonnet 4       & \textbf{83.9\%} & \underline{80.4\%} & 82.1\% & 82.1\% & 82.1\% \\
Claude Sonnet 4$^+$   & \textbf{76.8\%} & 75.0\% & 73.2\% & 75.0\% & \underline{71.4\%} \\
\midrule
\textit{Average}      & \textit{68.3\%} & \textit{67.5\%} & \textit{67.5\%} & \textit{\textbf{68.5\%}} & \textit{\underline{67.3\%}} \\
\bottomrule
\end{tabular}
\caption{Pass rates by model and condition.}
\label{tab:main-results}
\floatnote{Each cell aggregates 56 tasks with three repetitions by majority vote. Bold indicates the best condition in each row; underline indicates the worst. Across model-condition cells, task-level standard deviations range from 0.37 to 0.50.}
\end{table*}

Table~\ref{tab:main-results} presents pass rates by model and condition. The aggregate spread across all five conditions is just 1.2~pp (67.3\% to 68.5\%). The mixed-effects model (Table~\ref{tab:glmm}) gives the same conclusion: no condition reaches significance at $\alpha$=0.05, and all four skill conditions fall well short (all $p$$\geq$0.396).

Paired McNemar tests tell a similar story: for each skill condition, skills help and hurt on roughly equal numbers of task$\times$model pairs (e.g., Core+Examples: 20 helped vs.\ 19 hurt, $p$=1.000). Full skills consume ${\sim}4.5{\times}$ the input tokens of No-Skill (mean 1{,}293 vs.\ 287 input tokens), so the tested knowledge-injection mechanism adds cost without measurable accuracy gain.

\begin{table}[ht]
\centering
\small
\begin{tabular}{@{}lrr@{}}
\toprule
\textbf{Condition} & \textbf{$\hat{\beta}$} & \textbf{$p$} \\
\midrule
Full         & $-$0.008 & 0.497 \\
Core-Only    & $-$0.008 & 0.497 \\
Core+Ex      & $+$0.002 & 0.865 \\
Core+Refs    & $-$0.010 & 0.396 \\
\bottomrule
\end{tabular}
\caption{Mixed-model condition effects.}
\label{tab:glmm}
\floatnote{$n$=2{,}520 majority-vote outcomes; No-Skill is the reference. $\hat{\beta}$ is the effect in pass-rate points. The model includes task random intercepts and a model random-effect variance component. Logistic GEE and permutation tests yield substantively identical conclusions (all $p$$>$0.45).}
\end{table}

\subsection{Task Source Does Not Explain the Null}

Because half of the tasks are self-authored and half are externally grounded, we check whether the null result is an artifact of task construction. Table~\ref{tab:source-split} shows that the source split does not change the conclusion. Full skills are slightly below No-Skill for both self-authored tasks ($-$0.8~pp, 95\% CI [$-$5.2, $+$3.2]) and externally grounded tasks ($-$0.8~pp, 95\% CI [$-$5.2, $+$2.8]). Averaging across all four skill conditions yields the same pattern: $-$0.7~pp on self-authored tasks and $-$0.5~pp on externally grounded tasks. Thus, the aggregate null is not driven by one source group.

\begin{table}[ht]
\centering
\footnotesize
\resizebox{\columnwidth}{!}{%
\begin{tabular}{@{}lrrrrr@{}}
\toprule
\textbf{Task Source} & \textbf{Tasks} & \textbf{No-Skill} & \textbf{Full} & \textbf{Avg Skill} & \textbf{Skill Diff.} \\
\midrule
Self-authored & 28 & 70.2 & 69.4 & 69.5 & $-$0.7 [$-$4.3, $+$2.6] \\
Externally grounded & 28 & 66.3 & 65.5 & 65.8 & $-$0.5 [$-$4.3, $+$2.7] \\
\bottomrule
\end{tabular}%
}
\caption{Task-source split.}
\label{tab:source-split}
\floatnote{Entries are percentages based on majority-vote outcomes. Avg Skill averages Full, Core-Only, Core+Examples, and Core+Refs. Skill Diff. reports Avg Skill minus No-Skill, with 95\% bootstrap CIs over task-level paired differences.}
\end{table}

\subsection{Token-Matched Control}

We additionally compare Full skills against a token-matched Length-Control condition containing task-irrelevant office-supply guidance. This analysis helps separate the effect of generated data-science content from the effect of adding similarly long, skill-formatted context. Length-Control reaches 66.9\% pass rate, compared with 67.5\% for Full and 68.3\% for No-Skill. The paired comparisons in Table~\ref{tab:length-control-results} provide no evidence that the data-science skill content improves over the matched control: Full exceeds Length-Control by only $+$0.6~pp (95\% CI [$-$2.2, $+$3.4], $p$=0.775).

Length-Control also does not show strong evidence of harm relative to No-Skill ($-$1.4~pp, 95\% CI [$-$3.4, $+$0.6], $p$=0.311), although the point estimate is negative. Thus, the data do not support a simple prompt-length explanation for the null, nor is there measurable evidence that the generated data-science content adds useful information beyond a similarly sized, skill-formatted control. We treat this as a control analysis, not as proof that irrelevant content has no effect.

\begin{table}[ht]
\centering
\footnotesize
\resizebox{\columnwidth}{!}{%
\begin{tabular}{@{}lrrrc@{}}
\toprule
\textbf{Comparison} & \textbf{Diff.} & \textbf{95\% CI} & \textbf{Help/Hurt} & \textbf{$p$} \\
\midrule
Length-Control vs.\ No-Skill & $-$1.4 pp & [$-$3.4, $+$0.6] & 14/21 & 0.311 \\
Full vs.\ Length-Control     & $+$0.6 pp & [$-$2.2, $+$3.4] & 26/23 & 0.775 \\
Full vs.\ No-Skill           & $-$0.8 pp & [$-$3.8, $+$2.0] & 19/23 & 0.644 \\
\bottomrule
\end{tabular}%
}
\caption{Paired length-control comparisons.}
\label{tab:length-control-results}
\floatnote{Each condition has 504 majority-vote outcomes. Differences are percentage-point changes for the left condition relative to the right. Help/Hurt counts task$\times$model pairs where the left condition changes a failure to a pass versus a pass to a failure. Confidence intervals bootstrap over task-level paired differences; $p$ values use two-sided McNemar tests.}
\end{table}

\subsection{Prompt-Level Priority Control}

We next test whether a prompt-level task-over-skill directive mitigates the suspected skill-task conflicts. Full+Priority reaches 68.7\% pass rate, compared with 67.5\% for Full and 68.3\% for No-Skill. The paired difference is small and not statistically reliable: Full+Priority exceeds Full by $+$1.2~pp (95\% CI [$-$1.0, $+$3.6], $p$=0.362) and exceeds No-Skill by $+$0.4~pp (95\% CI [$-$2.4, $+$3.0], $p$=0.875). In the mixed-effects model, the Full+Priority coefficient relative to No-Skill is $+$0.004 ($p$=0.744).

The cell transitions suggest a tradeoff rather than a clean repair. Full+Priority recovers 12 of the 23 task$\times$model cells where No-Skill passes but Full fails, consistent with the directive resolving some skill-task conflicts. But it also introduces 8 new failures among cells where both No-Skill and Full pass, and loses 4 cells where Full alone helped. The added failures are mostly format or output-discipline errors, suggesting that the priority rule can pull the model back toward the task while weakening useful discipline supplied by the skill.

The directive also does not materially repair the main Data Preparation harm. Data Preparation rises from 50.8\% under Full to 51.6\% under Full+Priority, still below the 57.1\% No-Skill baseline. A simple user-message priority sentence is therefore not enough to remove the conflict-heavy failure mode. Stronger mechanisms, such as system-message placement, retrieval-gated skill snippets, or schema/tool-level constraints, remain open.

\subsection{No Component Shows a Reliable Benefit}

The aggregate null could still hide component-level cancellation: one section might help while another hurts. The component-level comparisons are small and non-significant, but they help characterize why the aggregate effect is flat:

\begin{itemize}
    \item \textbf{Reference Notes.} Core+Refs (67.3\%) falls below No-Skill (68.3\%) by $-$1.0~pp, and below Core-Only (67.5\%) by $-$0.2~pp.
    \item \textbf{Worked Examples.} Core+Examples (68.5\%) is closest to No-Skill (68.3\%), differing by just $+$0.2~pp.
    \item \textbf{Full Skill.} Full (67.5\%) matches Core-Only, suggesting that adding both Examples and Reference Notes to the base procedure produces no net change.
\end{itemize}

Because the design crosses Worked Examples and Reference Notes, we can also estimate their interaction. Adding Reference Notes on top of Core+Examples changes performance by $-$1.0~pp, close to the $-$0.2~pp change from adding Reference Notes on top of Core-Only, yielding an interaction of $-$0.8~pp. This small value gives little evidence that a large positive effect from examples is being canceled by a large negative effect from reference notes. It does not, however, reveal whether models internally ignore, give less weight to, or reconcile competing sections.

These near-zero differences are not just cancellation across models: 4 of 9 models achieve their best pass rate under No-Skill, and no single skill condition is best for more than 3 models. The component ablation therefore gives little evidence that any section helps on its own.

\subsection{Failures Vary by Workflow Difficulty}

The ablation results provide little support for the simplest cancellation story, but they do not by themselves explain why skill content fails in different workflows. Table~\ref{tab:family-results} shows that the aggregate null is not a single uniform failure. Baseline difficulty differs sharply across lifecycle stages, and each regime points to a different plausible reason why flat, LLM-generated skill injection may fail to help.

\begin{table}[ht]
\centering
\footnotesize
\resizebox{\columnwidth}{!}{%
\begin{tabular}{@{}lrrrrrrl@{}}
\toprule
\textbf{Lifecycle Stage} & \textbf{No-Skill} & \textbf{Full} & \textbf{L-Control} & \textbf{C-Only} & \textbf{C+Ex} & \textbf{C+Refs} & \textbf{Range (SD)} \\
\midrule
Data Extraction    & 88.9 & 91.3 & 89.7 & 89.7 & 90.5 & 89.7 & 0 to 100 (26.1) \\
Reporting & 97.6 & 99.2 & 97.6 & 99.2 & 98.4 & 99.2 & 78 to 100 (6.4) \\
Data Preparation & 57.1 & 50.8 & 52.4 & 52.4 & 54.8 & 52.4 & 0 to 100 (32.6) \\
Stat.\ Analysis & 29.4 & 28.6 & 27.8 & 28.6 & 30.2 & 27.8 & 0 to 67 (28.4) \\
\bottomrule
\end{tabular}%
}
\caption{Pass rates by lifecycle stage.}
\label{tab:family-results}
\floatnote{Entries are percentages. L-Control is the token-matched task-irrelevant control. Range and SD summarize within-stage task difficulty under No-Skill.}
\end{table}

\begin{figure}[t]
\centering
\includegraphics[width=\columnwidth]{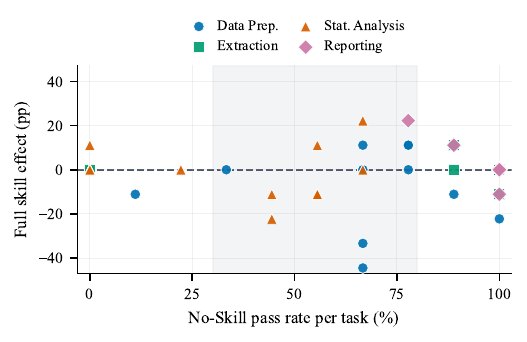}
\Description{Task-level scatter plot of No-Skill pass rate against Full skill effect. Many tasks lie near zero effect, with ceiling and floor clusters visible.}
\caption{Full-skill effects by task.}
\label{fig:task-effects}
\floatnote{Each point is one task, averaged over nine models after majority vote. The shaded region marks the 30 to 80\% No-Skill range used for the informative-task analysis.}
\end{figure}

\noindent Figure~\ref{fig:task-effects} shows the same pattern at task level. Many tasks sit near ceiling or floor under No-Skill, limiting the room for measurable gains. Within the informative 30 to 80\% band, full-skill effects remain mixed and often negative, especially for Data Preparation and Statistical Analysis tasks.

\textbf{Pattern 1: Redundancy at ceiling.} Data extraction (88.9\%) and reporting (97.6\%) exhibit near-ceiling baselines, leaving little room for improvement. This may reflect both the relative simplicity of many tasks in these stages and the extensive representation of SQL and JSON generation in LLM training corpora. In these stages, skill content appears mostly redundant with behavior the model already has in this benchmark.

\textbf{Pattern 2: Conflicting heuristics at moderate baselines.} Data preparation (57.1\%) sits in the informative difficulty range, yet skills show a slight negative direction ($-$2.3 to $-$6.3~pp). Examining condition-flip cases suggests a plausible explanation: among the 14 task$\times$model pairs where Core-Only passes but Core+Refs fails, data preparation tasks account for 8 of 14 flips. The remaining flips span Statistical Analysis (3), SQL (2), and Reporting (1), so the example below illustrates the largest category rather than all failures. We treat this analysis as diagnostic evidence rather than a causal mechanism claim. One representative case illustrates the problem:

\begin{quote}
\small
\textbf{Task instruction}: ``fill missing ages with \emph{median}, missing ports with `Unknown'.'' \textbf{Reference Notes}: ``For numeric columns with $<$30\% missing, apply \emph{forward-fill}. For string columns, fill with `UNKNOWN'.'' Under Core-Only, the output follows the task and passes. Under Core+Refs, the output instead follows the skill's generic heuristics, producing output that fails verification.
\end{quote}

When LLM-generated skills contain generic heuristics that conflict with task-specific instructions, outputs sometimes follow the skill content, introducing errors that would not occur without it. The risk is not just low skill quality: plausible but over-general rules can compete with the actual task.

\textbf{Pattern 3: Capability gaps at floor.} Statistical analysis tasks (29.4\%) require multi-step reasoning over hypothesis tests and effect-size computations, and the low baseline suggests that models often struggle regardless of skill content. The best skill condition adds only $+$0.8~pp, which is consistent with a computational or statistical reliability bottleneck that prompt-level knowledge injection alone does not fix.

Difficulty also varies substantially within each lifecycle stage. Of the 56 tasks, 17 fall in the informative 30 to 80\% No-Skill baseline range (9 Data Preparation and 7 Statistical Analysis tasks, plus 1 Reporting task). The remaining 39 tasks are near ceiling (28 tasks above 80\%) or near floor (11 tasks below 30\%), so the aggregate average should be read together with the stage-level results.

On the 17 informative tasks, all four skill conditions show a slight negative direction, but all intervals include zero: Full $-$2.6~pp (95\% CI [$-$11.1, $+$5.2]), Core-Only $-$5.2~pp ([$-$15.7, $+$4.6]), Core+Examples $-$2.0~pp ([$-$7.8, $+$3.3]), and Core+Refs $-$4.6~pp ([$-$12.4, $+$2.6]). This qualifies the negative pattern: the informative subset is the right place to look for gains, but the current sample supports only a diagnostic conclusion, not a precise small-effect estimate.

\subsection{Skills Do Not Interact with Model Group}

\begin{figure}[t]
\centering
\includegraphics[width=\columnwidth]{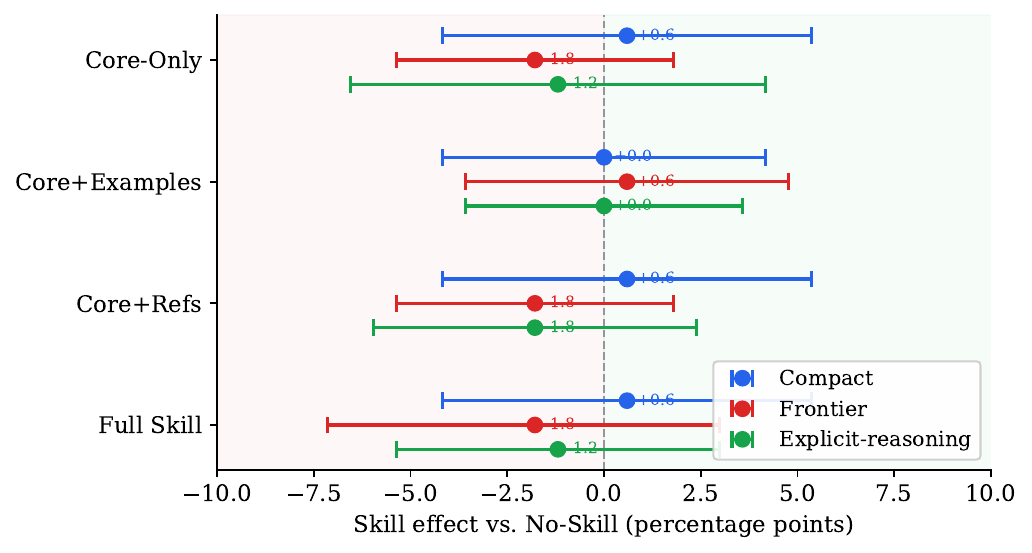}
\Description{Bootstrap confidence intervals for skill effects by model group, showing all intervals include zero across compact, frontier, and explicit-reasoning groups.}
\caption{Skill effects by model group.}
\label{fig:tier-effects}
\floatnote{Effects are relative to No-Skill with 95\% bootstrap CIs. No interval excludes zero.}
\end{figure}

One natural hypothesis is that the aggregate null conceals opposing effects across model groups: perhaps compact models benefit from structured guidance while frontier models are harmed. Figure~\ref{fig:tier-effects} shows stratified bootstrap results by group: compact (GPT-4o-mini, Gemini Flash, Claude Haiku), frontier default (GPT-4o, Gemini Pro, Claude Sonnet), and explicit-reasoning (o3-mini, Gemini Flash$^+$, Claude Sonnet$^+$). The intervals include zero across all three groups, providing no reliable evidence of a group-specific benefit at the current sample size.

Compact models show effects centered near zero, ranging from 0.0 to $+$0.6~pp across conditions. Frontier default models show effects from $-$1.8 to $+$0.6~pp, with Full, Core-Only, and Core+Refs tied at $-$1.8~pp. Explicit-reasoning configurations show a similar pattern, with the largest point estimate at $-$1.8~pp for Core+Refs. All intervals include zero.

Individual models within each group vary substantially: Gemini Flash gains $+$7.1~pp from Core+Examples while GPT-4o-mini loses $-$5.4~pp from the same condition. This heterogeneity does not form a systematic group-level pattern. Skills are not reliably more useful for weaker, stronger, or explicit-reasoning configurations. Appendix~\ref{app:stage-tier} reports a stage-by-group diagnostic table; it does not reveal a hidden explicit-reasoning gain on data preparation or statistical analysis.

\subsection{Token Cost Without Accuracy Gain}

Full-skill injection increases input length by about $4.5{\times}$ across providers (Table~\ref{tab:overhead}). Because providers charge for input tokens, this overhead increases cost before any change in output length. Input-token counts are deterministic for a fixed prompt and tokenizer, so they provide a stable measure of prompt-side overhead. We report this table for non-thinking configurations because explicit reasoning modes add provider-specific reasoning tokens, which would conflate skill-injection overhead with reasoning-budget overhead.

\begin{table}[ht]
\centering
\small
\resizebox{\columnwidth}{!}{%
\begin{tabular}{@{}llrrr@{}}
\toprule
\textbf{Provider} & \textbf{Condition} & \textbf{Mean Input Tok} & \textbf{P50 Output Tok} & \textbf{P50 Total Tok} \\
\midrule
\multirow{2}{*}{OpenAI}    & No-Skill & 263  & 337 & 598 \\
                           & Full     & 1{,}192 & 353 & 1{,}502 \\
\midrule
\multirow{2}{*}{Google}    & No-Skill & 303  & 174 & 500 \\
                           & Full     & 1{,}335 & 203 & 1{,}687 \\
\midrule
\multirow{2}{*}{Anthropic} & No-Skill & 296  & 432 & 736 \\
                           & Full     & 1{,}354 & 783 & 2{,}101 \\
\bottomrule
\end{tabular}%
}
\caption{Token footprint under No-Skill and Full for non-thinking configurations.}
\label{tab:overhead}
\floatnote{Mean input tokens average deterministic prompt token counts over tasks and listed models. Output and total tokens use medians over runs. OpenAI pools GPT-4o and GPT-4o-mini; Google uses Gemini 2.5 Flash with thinking disabled; Anthropic pools Claude Haiku 4.5 and Claude Sonnet 4 with thinking disabled.}
\end{table}

Output length changes are less uniform. Median output changes only modestly for OpenAI and Google, but rises more sharply for Anthropic in these non-thinking configurations. Thus, the total token cost of skill injection depends on both the provider and the model's response behavior.

Because the aggregate accuracy effect is null, this additional token cost comes without a measured accuracy gain in our setting. We do not report latency as a precise overhead estimate because runs were not interleaved or controlled for time-of-day and API-load variation.

\section{Implications for Data-Science Agent Design}

For the setting we test, the design message is limited but useful: single-shot, flat-file, LLM-generated skill packages are not a reliable default. Different lifecycle stages appear to need different kinds of support.

\textbf{Lifecycle-dependent strategy.} The three failure patterns suggest that a one-size-fits-all knowledge injection strategy may be brittle across the data-science lifecycle. For high-baseline stages (data extraction, reporting), skills add cost ($\sim$4.5$\times$ input-token overhead) with little observed room for improvement. For moderate-baseline stages (data preparation), generic skills can harm performance by introducing conflicting heuristics. For low-baseline stages (statistical analysis), the results are consistent with a computation and statistical-reasoning bottleneck, so prompt-level instructions alone may be insufficient.

\textbf{Task specificity over family coverage.} Our skills target entire lifecycle stages rather than individual tasks. SkillsBench~\cite{skillsbench2026} found that focused skills with two to three modules outperform comprehensive documentation; our family-level skills are closer to the comprehensive end. The conflicting-heuristics failure at the data-preparation stage is consistent with this breadth: a single skill cannot anticipate the particular imputation method each task requires. Task-specific skills that match exact requirements may avoid this failure mode, though at the cost of scalability.

\textbf{Selective loading and priority.} Our experiment injects all skill content as a single flat file. Anthropic's recommended architecture~\cite{anthropicskills2025} instead stores reference materials in separate files loaded only on demand. Because our conflicting-heuristics failures are disproportionately associated with Reference Notes ($-$1.0~pp vs.\ No-Skill), reference material should be gated rather than always prepended when systems support selective loading. Full+Priority also suggests that a user-message priority sentence is too weak; stronger priority mechanisms belong in system instructions, retrieval policy, or schema and tool constraints.

\textbf{Feedback over front-loaded guidance.} For workflows that require statistical computation or complex data cleaning, execution feedback may matter more than front-loaded generic guidance. The failure patterns point toward agents that retrieve narrower guidance, execute code, validate outputs, and revise, rather than agents that paste a full family-level skill into the first prompt.

\section{Limitations}

Our study has three main limitations. First, the scope is narrow. We evaluate 56 single-turn tasks in one domain, and only 17 tasks fall in the informative 30 to 80\% baseline range. The 95\% bootstrap CIs span $\pm$4 to 6~pp, so the study can rule out moderate average effects but not small effects (1 to 3~pp). We also do not include an expert-written or task-specific positive-control skill; that choice matches the low-curation workflow we test, but leaves the benchmark's sensitivity to carefully authored skills unmeasured.

Second, the setting is not a full data-science agent. It omits multi-turn planning, tool use, code execution, memory, iterative revision, and upstream product-framing tasks that require business context or stakeholder judgment. Skill placement is also fixed: skills are prepended as user-message content, with no system-level priority rule, retrieval gate, or schema/tool constraint. Our findings therefore apply most directly to single-shot, flat-file, LLM-generated, family-level skills on deterministically verifiable execution tasks.

Third, several confounds remain. Skill conditions increase prompt length (from a mean of 287 input tokens for No-Skill to 1{,}293 for Full). The Length-Control condition reduces concern that length alone explains the null, but it uses one irrelevant domain, office-supply setup, so it cannot characterize all forms of unrelated context. The use of one generated skill per lifecycle stage also ties section effects to four skill instances. Sampling multiple generated skills per stage would estimate inter-generation variance, but would no longer match the one-candidate low-curation workflow we study.

There is also an authorship overlap. Gemini 2.5 Pro assisted with skill writing and with drafting the self-authored task prompts and initial gold outputs, and Gemini 2.5 Pro is one of the evaluated model configurations. Deterministic validation, fixed gold files, and within-task condition comparisons reduce the risk that scoring directly rewards Gemini outputs. Still, the overlap could affect task style, formatting conventions, or absolute difficulty, especially for exact-match tasks. A cleaner design would separate task authoring, skill authoring, and model evaluation.

Finally, the condition-flip review gives only partial evidence about mechanism. It can identify overt cases where the output follows a generic skill heuristic over the task instruction, but it cannot count all cases where the skill was ignored, partially used, or internally reconciled with the task. We therefore report the flip analysis as diagnostic rather than as a complete instruction-adherence taxonomy.

\section{Conclusion}

We evaluated whether low-curation, LLM-generated skills improve data-science agent performance across four data-science lifecycle stages when skills are injected directly as flat prompt content. Across 56 tasks, nine model configurations, and three providers, neither the full generated skill nor any ablated skill variant significantly outperforms the task-only baseline. A token-matched control also performs similarly to the full skill, and a prompt-level task-over-skill directive does not reliably improve performance. The null result is consistent across skill components and model groups, and it aligns with three diagnostic patterns: redundancy at high baselines, conflicting heuristics at moderate baselines, and statistical-reasoning bottlenecks at low baselines.

For practitioners building data-science agents, these results support a cautious default: LLM-generated, family-level skill injection did not improve performance in this single-shot setting. The next tests should compare task-specific skills, selective loading, system-level priority rules, and execution-feedback agents.

\bibliography{references}

\clearpage
\appendix

\section{Length-Control Token Matching}
\label{app:length-control}

The Length-Control condition uses a task-irrelevant office-supply setup skill for each lifecycle stage. The control skills contain coherent procedural content about arranging a shared supply station, including activation guidance, steps, examples, and reference notes. They preserve the same flat injection wrapper and skill-like markdown organization as the Full skills, but remove data-science content. Token counts were computed with \texttt{tiktoken} v0.13.0 under both \texttt{o200k\_base} and \texttt{cl100k\_base}. Delta is computed as $(\text{Length-Control} - \text{Full}) / \text{Full}$.

\begin{table}[ht]
\centering
\small
\resizebox{\columnwidth}{!}{%
\begin{tabular}{@{}lrrr rrr@{}}
\toprule
& \multicolumn{3}{c}{\texttt{o200k\_base}} & \multicolumn{3}{c}{\texttt{cl100k\_base}} \\
\cmidrule(lr){2-4}\cmidrule(lr){5-7}
\textbf{Lifecycle Stage} & \textbf{Full} & \textbf{Control} & \textbf{Delta} & \textbf{Full} & \textbf{Control} & \textbf{Delta} \\
\midrule
Data Preparation      & 725  & 721  & $-0.55\%$ & 718  & 724  & $+0.84\%$ \\
Data Extraction       & 910  & 902  & $-0.88\%$ & 912  & 908  & $-0.44\%$ \\
Statistical Analysis  & 1094 & 1090 & $-0.37\%$ & 1095 & 1098 & $+0.27\%$ \\
Reporting             & 947  & 942  & $-0.53\%$ & 947  & 953 & $+0.63\%$ \\
\bottomrule
\end{tabular}%
}
\caption{Token matching for the Length-Control skills.}
\label{tab:length-control-token-matching}
\floatnote{The control skills are token-matched to the corresponding Full skills within 1\% under both tokenizers.}
\end{table}

\section{Stage-by-Group Diagnostic Effects}
\label{app:stage-tier}

Table~\ref{tab:stage-tier} reports Full-minus-No-Skill effects within each lifecycle stage and model group. These are descriptive diagnostics, not a formal search over all stage, group, and component interactions. The table does not show evidence that explicit-reasoning configurations uniquely benefit from Full skills on statistical analysis or data preparation; those two rows are near zero or negative across groups, with wide confidence intervals.

\begin{table}[ht]
\centering
\scriptsize
\resizebox{\columnwidth}{!}{%
\begin{tabular}{@{}lccc@{}}
\toprule
\textbf{Lifecycle Stage} & \textbf{Compact} & \textbf{Frontier} & \textbf{Explicit-reasoning} \\
\midrule
Data Preparation & $-$4.8 [$-$16.7, $+$7.1] & $-$7.1 [$-$19.0, $+$2.4] & $-$7.1 [$-$19.0, $+$2.4] \\
Data Extraction & $+$4.8 [0.0, $+$11.9] & $+$0.0 [$-$7.1, $+$7.1] & $+$2.4 [0.0, $+$7.1] \\
Stat.\ Analysis & $+$0.0 [$-$9.5, $+$9.5] & $+$0.0 [$-$14.3, $+$14.3] & $-$2.4 [$-$9.5, $+$4.8] \\
Reporting & $+$2.4 [$-$7.1, $+$14.3] & $+$0.0 [0.0, 0.0] & $+$2.4 [0.0, $+$7.1] \\
\bottomrule
\end{tabular}%
}
\caption{Full-skill effects by lifecycle stage and model group.}
\label{tab:stage-tier}
\floatnote{Entries are percentage-point differences relative to No-Skill with 95\% bootstrap CIs over task-level paired differences.}
\end{table}

\end{document}